\documentclass[a4paper, 10 pt, conference]{ieeeconf}  

\IEEEoverridecommandlockouts                              

\usepackage{graphics} 
\usepackage{epsfig} 
\usepackage{amsmath} 
\usepackage{amssymb}  
\usepackage{multirow}
\usepackage{subfig}
\usepackage{hyperref}

\usepackage{svg} 

\title{\LARGE \bf
Towards Robust Deep Reinforcement Learning for Traffic Signal Control: Demand Surges, Incidents and Sensor Failures 
}

\author{Filipe Rodrigues$^{1}$ and Carlos Lima Azevedo$^{1}$
\thanks{$^{1}$Technical University of Denmark (DTU), Bygning 116B, 2800 Kgs. Lyngby, Denmark.
        Emails: {\tt\small rodr@dtu.dk, climaz@dtu.dk}}%
}

\begin{document}

\maketitle
\thispagestyle{empty}
\pagestyle{empty}

\begin{abstract}

Reinforcement learning (RL) constitutes a promising solution for alleviating the problem of traffic congestion. In particular, deep RL algorithms have been shown to produce adaptive traffic signal controllers that outperform conventional systems. However, in order to be reliable in highly dynamic urban areas, such controllers need to be robust with the respect to a series of exogenous sources of uncertainty.

In this paper, we develop an open-source callback-based framework for promoting the flexible evaluation of different deep RL configurations under a traffic simulation environment. With this framework, we investigate how deep RL-based adaptive traffic controllers perform under different scenarios, namely under demand surges caused by special events, capacity reductions from incidents and sensor failures. We extract several key insights for the development of robust deep RL algorithms for traffic control and propose concrete designs to mitigate the impact of the considered exogenous uncertainties. 
\end{abstract}

\section{INTRODUCTION}

Traffic congestion is a long-standing major issue with critical consequences at the environmental, social and economic levels. For example, according to a recent study from INRIX \cite{inrix2019}, in 2018, Americans lost an yearly average of 97 hours due to congestion, costing nearly \$87 billion - an average of \$1,348 per driver. Since a significant portion of the traffic delays occurs at signalized intersections, developing efficient and adaptive traffic signal control systems is of the uttermost importance. Different control-theory based methods have been proposed and widely implemented in practice \cite{sims1980sydney, hunt1981scoot, gartner1983opac}. In recent years, several artificial intelligence techniques such as neural networks, fuzzy inference systems and reinforcement learning (RL) have also been proposed, and the exploration of the latter has seen significant progress lately \cite{yau2017survey,el2014design}. The fact that RL algorithms operate in a delayed return environment, where it can be difficult to understand which action leads to which outcome over many time steps, and the fact that RL agents can learn online in a dynamic environment, makes them perfectly suited for developing adaptive traffic signal control systems. More recently, the advances in deep learning and, specifically, in deep RL, have made the application of RL approaches for traffic signal control even more appealing, since they are capable of handling a large number of high-dimensional states by avoiding the discretization of the state space \cite{liang2018deep, schutera2018distributed}. 

Despite the recent rise of deep RL applications for traffic signal control, little work has been done regarding their robustness performance and uncertainty handling. de Weck et al. \cite{de2007classification} classified uncertainties in early system design as endogenous and exogenous. While the first can generally be influenced by the control system designer, the latter comes from the interaction of the system with the environment. Traffic control systems are also subject to both types of uncertainties. Endogenous uncertainties relate, for example, to the design of the reward function, the RL learning algorithm used, state and action space definition (see \mbox{e.g.} \cite{muresan2019adaptive}), etc. Such aspects have been studied to a great extent in the traffic control literature. For example, Seyed et al. \cite{mousavi2017traffic} compare two classes of RL algorithms: deep policy-gradient and value-function-based methods. Similarly, El-Tantawy et al. \cite{el2014design} provide an extensive comparison between different reward functions, state representations, actions spaces and exploration strategies. For a recent review of reinforcement learning approaches for traffic signal control, the interested reader is referred to \cite{yau2017survey}.

Exogenous uncertainties, however, have been studied to a much smaller extent. These may come from the demand, the supply or other components of the traffic signaling system itself. While it is known that some variations to \textit{an average day} traffic demand on a network are always expected, its full range of variations and its uncertain frequencies are not often explicitly considered nor handled during the system design, especially for heavy stress scenarios \cite{park1999traffic, tong2015stochastic}. Incidents or other road blockages may affect the supply performance of the system at stake, often leading to network effects through queue overflowing to upstream intersections \cite{shed2003stochastic}. Finally, from the components of the traffic signaling system, common sources of uncertainty relate to sensing and communication errors. Traffic control may rely on a few different technologies (e.g., loop detectors, cameras or RFID sensors) for its operation \cite{li2012reliable}. It is known that each of these can have different performance and failure probabilities depending on many different factors, such as weather, occlusions or computer vision classification or detection errors (cameras), communication failures, lack of maintenance, etc. \cite{rajagopal2007health}. All the aforementioned uncertainties can dramatically affect the performance of RL algorithms, thus making learning policies, that are robust to their effect, crucial \cite{morimoto2005robust,pinto2017robust}. Aslani et al. \cite{aslani2017adaptive} presented one of the few comprehensive evaluations of both discrete and continuous RL-based traffic signals in a realistic setting (Tehran's network) for road blockages from pedestrians crossing and double parking, for high demand from non-recurring congestion and for different levels of sensor noise. Yet, a robustness analysis for deep RL techniques, where the introduction of neural networks as function approximators changes the characteristics of the learning system quite dramatically \cite{mnih2015human}, is still to be explored. 

In this paper, we focus on the exogenous sources of uncertainty and investigate how they affect the performance of signal control systems based on deep RL. For this purpose, we develop an open-source callback-based framework for the flexible evaluation of different RL configurations under a traffic simulation environment. Using this framework, we study the effect of demand surges, incidents and sensor failures in the performance of deep RL algorithms, namely duelling deep Q-learning, for traffic signal control. Furthermore, we identify a set of initial guidelines towards the development of robust traffic signal control systems based on deep RL, whose effectiveness we demonstrate empirically thereby establishing a publicly-available benchmark for future research on this topic. 


\section{MODELLING FRAMEWORK}
In this section we present both the modelling framework of the deep RL-based traffic signal controller and the architecture of an open-source CAllback-based REinforcement Learning framework (CAREL) to test the proposed and future deep RL-based traffic control systems (source code is available at: \url{http://github.com/fmpr/CAREL}).

\subsection{Robust deep RL for traffic signal control}

The general idea of RL is that of an agent interacting with an environment by observing its current state at time $t$, $\textbf{s}_t \in \mathcal{S}$, taking an action $\textbf{a}_t \in \mathcal{A}$ according to some policy $\pi$, and observing delayed rewards $r(\textbf{s},\textbf{a})$, corresponding to the reward for having taken action $\textbf{a}$ at state $\textbf{s}$. The goal of RL is then to learn an optimal policy $\pi^*$ that maximizes the discounted cumulative reward $R_t = \sum_{j=t}^\infty \gamma^{(j-t)} \, r_j$, where $\gamma \in [0,1]$ is a discount factor that trades-off the importance of immediate and future rewards \cite{sutton1998introduction}. 

The ability of RL algorithms to correlate immediate actions with the delayed returns that they produce makes them particularly well suited for developing policies for adaptive traffic controllers that try to optimize some indicator of intersection/network performance. A particularly popular class of RL algorithms that has been proven successful for traffic signal control \cite{liang2018deep,yau2017survey,el2014design} is Q-Learning. Q-Learning \cite{sutton1998introduction} seeks to learn an optimal policy $\pi^*$ implicitly by estimating the action-value function (or ``Q-function") that returns the expected future rewards by taking action $\textbf{a}_t$ given state $\textbf{s}_t$
\begin{align}
    Q^{\pi}(\textbf{s}_t,\textbf{a}_t) = \sum_{j=t}^T \mathbb{E}_{\pi}[r(\textbf{s}_j,\textbf{a}_j) | \textbf{s}_t, \textbf{a}_t],
\end{align}
and determining the best action at time $t$ by computing $\arg\max_{\textbf{a}_t} Q^{\pi}(\textbf{s}_t,\textbf{a}_t)$. However, conventional Q-learning algorithms can be very limiting for traffic control, since they assume the states $\textbf{s}_t$ to be discrete, thereby requiring a complicated process of discretization of real-valued traffic indicators (e.g. queue lengths and inductive loop data). Deep Q-learning \cite{mnih2013playing} relaxes this assumption by parameterizing the potentially extremely high-dimensional Q-function with a deep neural network $Q(\textbf{s},\textbf{a}; \boldsymbol\theta)$ with parameters $\boldsymbol\theta$ - \textit{deep Q-network} (DQN), thus allowing the expected future rewards for each action $\textbf{a}$ to be approximated by a complex non-linear function of a real-valued state vector $\textbf{s}$.

In order to improve stability and speed-up the convergence of DQN, a dueling architecture is used \cite{wang2015dueling}. In dueling DQN, the Q-function $Q^{\pi}(\textbf{s},\textbf{a}; \boldsymbol\theta)$ is further subdivided as the sum of a value function $V^{\pi}(\textbf{s}; \boldsymbol\theta)$, capturing the overall expected reward of being in state $\textbf{s}$ and taking probabilistic actions in future steps, and an advantage function $A^{\pi}(\textbf{s},\textbf{a}; \boldsymbol\theta)$, representing the advantage of choosing a particular action $\textbf{a}$ when in state $\textbf{s}$. The idea of dueling DQN is then to have a shared neural network architecture outputting both $V^{\pi}(\textbf{s}; \boldsymbol\theta)$ and $A^{\pi}(\textbf{s},\textbf{a}; \boldsymbol\theta)$ separately and estimating the Q-function as
\begin{align}
    Q^{\pi}(\textbf{s},\textbf{a}) = V^{\pi}(\textbf{s}; \boldsymbol\theta) + \bigg( A^{\pi}(\textbf{s},\textbf{a}; \boldsymbol\theta) - \frac{1}{|\mathcal{A}|} \sum_{a'} A^{\pi}(\textbf{s},\textbf{a}'; \boldsymbol\theta) \bigg),\nonumber
\end{align}
where subtracting the mean advantage has the benefit of increasing the stability of the optimization. Lastly, as a way to help mitigate the overoptimistic value estimation problem, a target network is also used \cite{mnih2015human}. 

So far, the RL approach described above is rather generic. In order to use it for robust traffic signal control, one must define the states, actions, rewards and neural network architecture appropriately. 

\textbf{State space:} The state definition can play a crucial role in developing RL approaches that are robust to changes in demand, supply and sensor reliability. Furthermore, it is essential to rely on sensory data that can be easily available in a modern road network, so that the developed methodology can be readily applied in practice. Therefore, state representations based on cumulative delays \cite{el2014design} and vehicle-level positioning and speeds \cite{liang2018deep} were not considered. Instead, our state representation is based on queue lengths, which can be easily obtained from loop sensor- or camera-based algorithms (input-output or shock-wave models) in the sensing component of existing traffic control systems \cite{liu2009real, ma2017traffic}. Letting $q_l^t$ denote the number of queued vehicles in lane $l$ at time $t$, the state is the defined as the maximum queue length associated with each phase $p$:
\begin{align}
    s_p^t = \max_{l \in \mathcal{L}_p} q_l^t, \quad \forall p \in \{1,\dots,P\},
    \label{eq:state}
\end{align}
where $P$ is the number of phases and $\mathcal{L}_p$ denotes the set of lanes associated with all movements $m$ in phase $p$ (Fig. \ref{fig:intersection}). 

The advantages of the max operator in Eq.~\ref{eq:state} are two-fold. First, it allows for a reduction of the state space in intersections with various lanes associated each phase, thereby speeding-up learning of the RL agent. Secondly, it allows for increased robustness to sensor failures, since a missing queue length measurement for a given lane will be attenuated by the max operation, rather than facing the RL agent with what would likely be a totally unseen input state. In order to further increase the robustness of the RL algorithm, we further propose extending the state representation by also including the elapsed time since the last green signal for each of the $P$ phases. The intuition is that this extra input variables are not subjective to sensor failures and are able to provide the RL agent with auxiliary information that can help it balance out the green assignments in case of noisy sensory inputs, while also preventing green starvation. 

\textbf{Action space:} We consider 2 possible action spaces that are common in the literature: time-extension and phase-selection. In time-extension, the action space $\mathcal{A}$ has size $2 \times P + 1$, and it includes the possibility of keeping the current phase timings or increasing/decreasing the green duration for each phase $p$ by 5s, as proposed in \cite{liang2018deep}. The RL agent then decides what action $a \in \mathcal{A}$ to take at the end of each cycle, which allows for a variable-length cycle but always keeps the phase sequence unaltered. In phase-selection, we have that $\mathcal{A} = \{1,\dots,P\}$ and an action $a \in \mathcal{A}$ corresponds to a transition to the selected phase with a yellow-phase transition of 3s as proposed, for example, in \cite{schutera2018distributed}. In order to stabilize learning and obtain a more reliable reward signal, we only allow for actions at every 10s. This has the effect of ensuring that enough evidence is collected for evaluating the reward function, while also preventing extremely short phases in a variable phase sequence setting. 

\textbf{Reward function:} For the ease of deployment of the developed controllers, we aimed at a reward function that can also be easily obtained in real environments (unlike reward functions based on the number of stops, cumulative delays, stop time, etc.). For that purpose, the reward function used in our experiments is also based on queue lenghts \cite{el2014design}:
\begin{align}
    r^t = \sum_{p=1}^{P} \Big(\max_{l \in \mathcal{L}_p} q_l^t \Big)^2 - \sum_{p=1}^{P} \Big(\max_{l \in \mathcal{L}_p} q_l^{(t-1)} \Big)^2.
    \label{eq:reward}
\end{align}
We also experimented with other reward functions from the literature (see \cite{el2014design}) during our preliminary experiments; we consistently found the reward in Eq.~\ref{eq:reward} to be the best. 

\textbf{Neural network architecture:} The neural network architecture used consists of 2 fully-connected hidden layers with ReLU activations, and an output layer of size $P$ or $2 \times P + 1$, depending on whether the action space corresponds to phase-selection or time-extension, respectively. In order to prevent overfitting and increase the robustness of the controller, we further use Dropout \cite{srivastava2014dropout} between all densely connected layers. With Dropout, units (along with their connections) are randomly dropped from the neural network during training. As it turns out, this is a key ingredient for developing controllers that are robust to, for example, sensor failures (see Section~\ref{subsec:sensor_failures}). The intuition is that Dropout prevents units from co-adapting too much, thereby encouraging the network to learn multiple alternative representations that depend on different combinations of the inputs and then averaging over them in order to output a prediction (e.g. Q-values). Thus, the effect of sensor anomalies/failures can be averaged out, by leveraging information from the other sensors and the extended state representation with the elapsed times since the last green. Although it a standard tool for preventing overfitting, this paper explicitly studies the effect of Dropout on the robustness of the learned policies to sensor failures. 

\textbf{Learning strategy:} For the phase-selection approach, we use the Adam optimizer with a learning rate of $10^{-4}$ and the update rate of the target network $\tau = 0.005$. As for time-extension, the best results were obtained for a learning rate of $10^{-3}$ and $\tau = 0.01$. For both approaches, we use experience replay with a memory size of $10000$ observations. Exploration is done through an $\epsilon$-greedy policy, with annealing for the $\epsilon$ parameter between $0.5$ and $0.1$ over the first $1000$ steps.

\subsection{CAREL architecture}

\begin{figure}[t]
\centering
\includegraphics[scale=0.44]{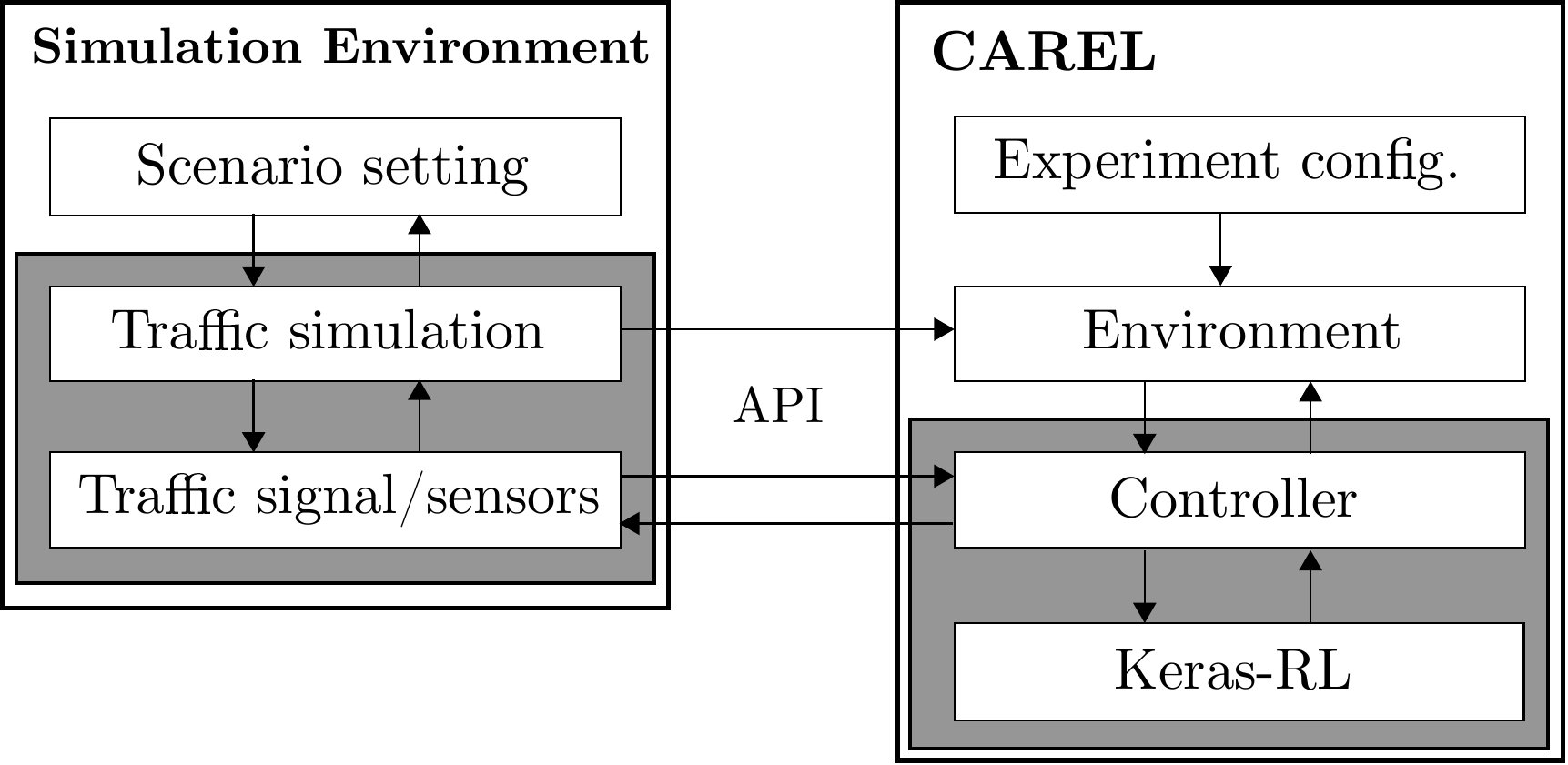}
\caption{CAREL framework.}
\label{fig:framework}
\vspace{-0.3cm}
\end{figure}

CAREL is structured into two main objects: the Environment and the Controller (Fig. \ref{fig:framework}). The first allows for the integration with the simulation environment being used through available APIs. It enables the transfer of simulation statistics for the overall experimental assessment and for the synchronous decision making between the controller and the simulation environment. The Controller handles the learning strategy, neural network architecture, reward function, action space and state space described above. In the current Python implementation, it relies on the Keras-RL library. 

With the proposed modular design, different control algorithms and configurations can be easily implemented in CAREL and transferred to other simulation or real environments. In this first deployment, the simulation environment used was AIMSUN Next \cite{casas2010traffic}. Although its use in traffic signal control assessment is available in the literature (e.g.: \cite{vilarinho2014capability}) only a couple of efforts related to deep RL-based traffic control have been reported: Casas \cite{casas2017deep} presented a deep deterministic policy gradient algorithm for the control of single intersections; and Gong et al. \cite{gong2019decentralized} proposed a network-level decentralized adaptive signal control based on double dueling deep Q-networks, testing it in an AM peak scenario for a suburban traffic corridor. However, no detailed provision of the integration architecture used was provided. Finally, in \cite{wu2017flow} an open-source computational framework called Flow is presented as a deep RL and control environment encapsulating traffic simulation software (including AIMSUN) for running and evaluation different automated-vehicle control scenarios. In Flow, the traffic simulation environment is called from within the deep RL environment, which has full control over the simulation environment. CAREL, however, is designed as a callback-based framework in which the simulation environment has full control over the simulation and only calls the RL-based controllers whenever an action is required or a new state has been observed. We argue that this provides a more efficient setting that is also closer to how reality works, thus making it easier to deploy the developed systems in real world environments. 

\section{EXPERIMENTS}

For the evaluation of the impacts of exogenous uncertainty on the proposed deep RL control, we considered three different dimensions: demand, supply and sensing. For each of these dimensions, a set of frequent scenarios in realistic settings were considered. The full space of uncertainty was not explored, but the analysis of the deep RL behaviour and performance is here at stake. Also, in this short paper we focus our analysis on a single intersection as a first step for understanding deep RL robustness. As in previous efforts for robustness analysis \cite{schutera2018distributed, aslani2017adaptive}, network and coordination can have significant impacts, but is left for future research.

\subsubsection{Network}
\begin{figure}[t]
\centering
\includegraphics[scale=0.38]{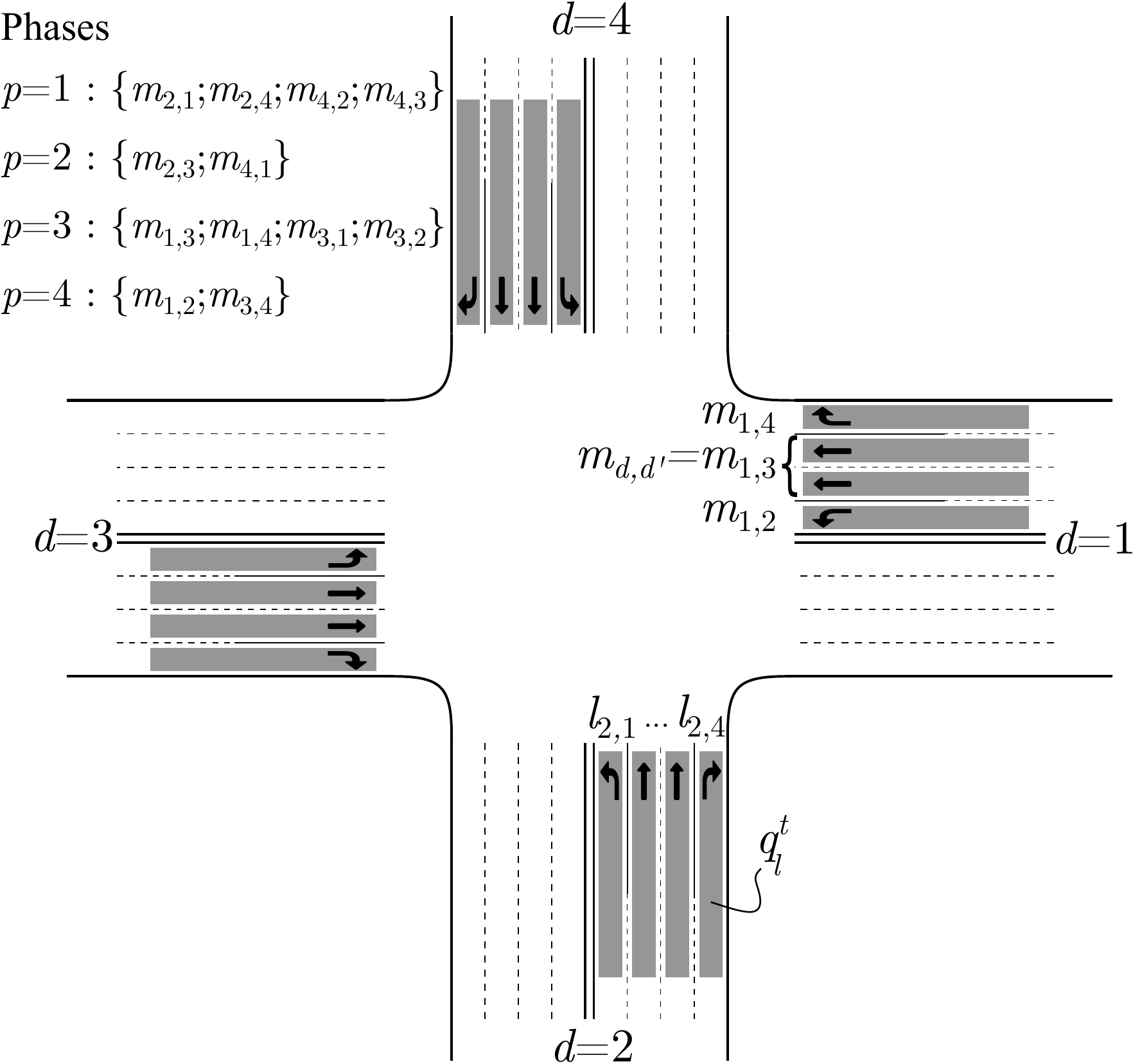}
\caption{Network layout.}
\label{fig:intersection}
\vspace{-0.0cm}
\end{figure}

A simple 4-leg intersection as represented in Fig. \ref{fig:intersection} was used in all experiments. Each leg or direction $d$ has 4 lanes, $l_{d,1...4}$, each with a single movement $m_{d,d'}$, where ${d,d'}$ represents the origin and destination direction. No side lanes were considered in the vicinity of the intersection. The queue length $q^{t}_{l}$ at time $t$ is available for each lane.

\subsubsection{Base scenario demand}
The demand for the base scenario was obtained from lane-based, 5 min loop sensor data collected by an existing adaptive traffic control system of a real intersection with the layout shown in Fig. \ref{fig:intersection}. The demand by 30 min for the 7:00 to 11:00 AM period for a weekday without disruption was used as reference (Table \ref{table:demand}).

\begin{table}[t]
\caption{Base scenario average demand}
\label{table:demand}
\begin{center}
\begin{tabular}{c|cccc}
 & $d=1$ & $d=2$ &$d=3$ & $d=4$ \\
\hline
$d=1$ & - & 109.7 (47.0) & 14.8 (6.1) & 12.8 (6.2) \\
$d=2$ & 167.4 (68.5) & - & 88.7 (35.3) & 219.2 (84.9) \\
$d=3$ & 89.3 (36.1) & 66.3 (27.7) & - & 152.6 (62.5)\\
$d=4$ & 32.2 (15.5) & 64.2 (27.1) & 154.2 (71.0) & - \\
\end{tabular}
\end{center}
\vspace{-0.5cm}
\end{table}

\subsubsection{Traffic Model}
AIMSUN \cite{casas2010traffic} was used in current CAREL's implementation. It includes a microscopic driver behaviour model, an explicit reproduction of different traditional traffic control systems and a flexible API. AIMSUN's car-following model is based on Gipps \cite{gipps1981behavioural}, where each driver sets limits to desired braking and acceleration rates and acceleration actions are ultimately influenced by vehicle characteristics, drivers reaction time and leader's trajectory. Vehicles can also be forced to slow down as they pass an incident to simulate the observed behaviour of drivers when passing the incident.
At the intersection level, AIMSUN allows for four types of traffic signal  control: unsignalised, pre-timed, actuated and external. For actuated control, a set of traditional configurations (namely minimum green, rest in red, allowance gap, passage gap) or even multi-ring or preemption configurations can be defined. For external controllers AIMSUN's API allows external changes to cycle lengths, green time durations and phase selections at each traffic light's stage. Finally, AIMSUN  provides a large number of performance measures, replicating existing traffic sensing capabilities and many global performance measures usually unobservable but relevant for optimality check.

\subsubsection{Baseline Controls}
In our experiments we considered two existing baseline controllers for comparison with the deep RL-based controllers: fixed timings with optimized parameters; and actuated with a dual-ring setup and AIMSUN's default parameters \cite{AimsunManual}.
For appropriate benchmark all controls relied on $p=1,...,4$ phases  presented in Fig. \ref{fig:intersection}.

\subsection{Performance evaluation}

For each scenario considered, three performance indicators were obtained from the simulation environment: total travel time, total delay and total stop time, by 5s and for the entire experiment duration. Each experiment accounted for a maximum of 80 episodes, each of which lasted 4h including a 10 min simulation warm-up time. This maximum number of episodes allowed for the evaluation of scenario dynamics and the stability of the deep RL performances. Finally, each experiment was replicated 10 times to allow for the quantification of the the variability from both the simulation and the control algorithm intrinsic stochasticities.

\subsubsection{Computational Setting}
The processing of all controller processes within CAREL were substantially faster than AIMSUN simulations and thus suitable for real-time performance. Experiments were ran in parallel on a 16 core AMD Ryzen Threadripper 2950, 32 threads, 128 Gb of RAM under Linux.

\subsection{Demand surge experiments}

We begin by analyzing the effect of different levels of demand and demand surges in the performance of the proposed RL approach. We consider the following 3 scenarios: 

\textbf{Low demand:} This scenario corresponds to the OD demand of the base scenario reduced by 30\%. 

\textbf{Before event:} Special events are know to cause significant disruptions in transport systems \cite{rodrigues2017bayesian}. In order to simulate the potential effect of an event that is about to start in the north of the intersection, in this scenario the north-bound demand (\mbox{i.e.} turn movements $m_{1,4}$, $m_{2,4}$ and $m_{3,4}$) is multiplied by a factor of 2 during the 2nd and 3rd hours of the total 4-hour simulation period.

\textbf{After event:} This scenario is the reverse of the one above: to mimic the effect of the end of an event in a venue just north of the intersection, we multiply the incoming demand from the north direction ($d=4$) to all other directions by a factor of 2.5 for a period of 2 hours in the middle of episode. 

The first rows of Table~\ref{table:results1} show the obtained results. Let us start by analyzing the results for the base scenario, which is based on the real demand data. As expected, we can verify that the use of an actuated controller results in better performance indicators than relying on fixed times. However, while the deep RL approach based on time-extension is barely able to outperform fixed times, we can observe that the phase-selection version significantly outperforms all the traditional approaches. This confirms the findings of previous works regarding the advantages of deep RL for traffic signal control \cite{yau2017survey,el2014design}. However, it questions the use of time-extension approaches as proposed for example in \cite{liang2018deep}, which we could not make perform as well as the phase-selection approach regardless of the amount of hyper-parameter tuning. 

What is perhaps more interesting is comparing the performance of the different methods when the demand varies. For example, we can observe that, as the demand decreases overall (``Low demand" scenario), the gap in the performance of the different methods narrows, making the phase-selection approach perform only slightly better an average than the actuated controller. This is particularly important given that a wide majority of the works researching the application of RL for traffic signal control considers fixed demands often with some arbitrary rate values \cite{Nishi2018TrafficSC,schutera2018distributed}. 

\begin{figure}[t!]
\centering
\includegraphics[scale=0.57,trim={0 0.2cm 0 1.3cm},clip]{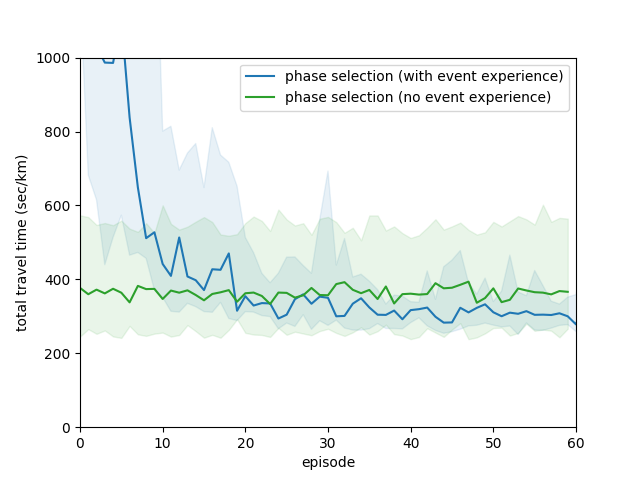}
\caption{Effect of event experience for ``After event" scenario.}
\label{fig:train_base_test_event1}
\vspace{-0.4cm}
\end{figure}

However, the effect of different demand levels only represents part of the problem. In order to learn RL-based controllers that are robust in lively urban environments, it is essential to account for demand surges like the ones caused by special events (\mbox{e.g.} music concerts, sports games, festivals or parades). In order to illustrate the importance of including scenarios demand surges during the training of the RL agent, Fig.~\ref{fig:train_base_test_event1} shows the result of directly deploying a RL-based controller that was learned on the base scenario on a scenario with demand surges (``After event" scenario), in comparison with the learning curve of a RL agent on a the latter scenario. Not only the average behaviour of the RL agent without experience in demand surges is significantly worse, but its performance is also very unstable, with cases of extremely poor travel times as evidenced by the 10 and 90 percentiles of the curves (shaded areas). 

\begin{figure*}
\subfloat[Neural network loss]{\includegraphics[scale=0.53,trim={0 0.2cm 0 1.3cm},clip]{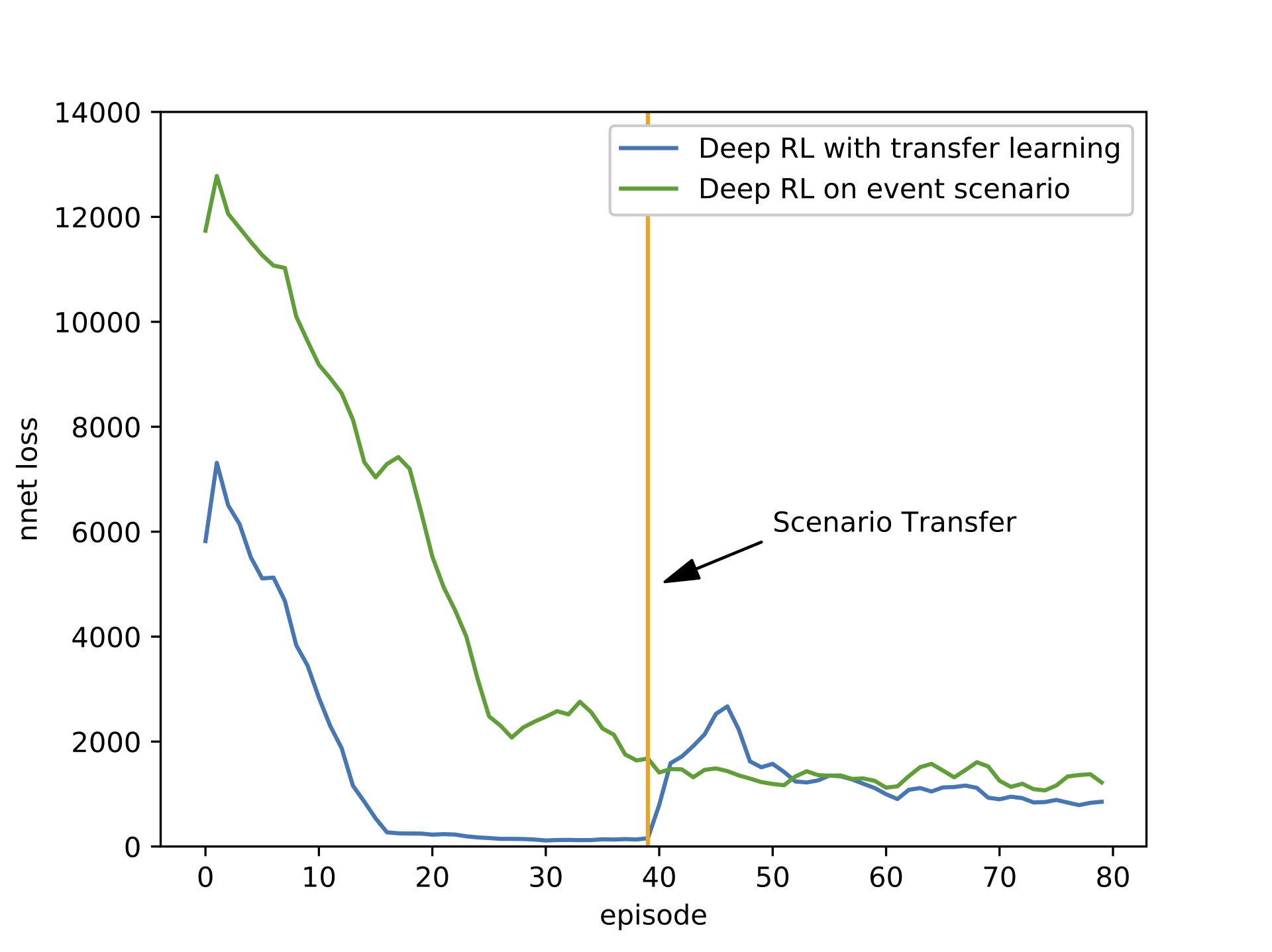}} 
\subfloat[Total travel time (sec/km)]{\includegraphics[scale=0.53,trim={0 0.2cm 0 1.3cm},clip]{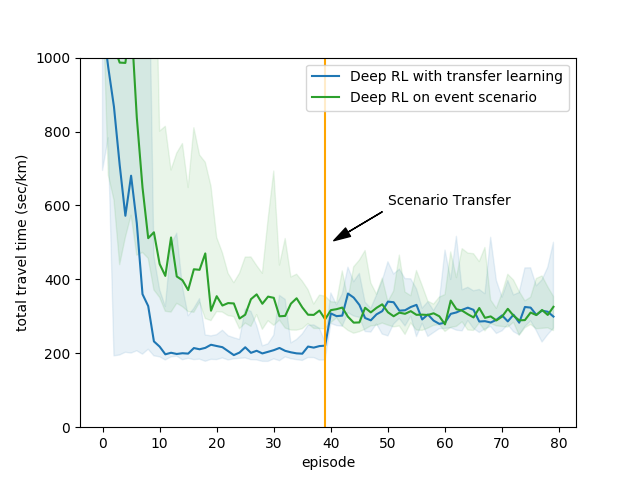} \label{subfig:transferb}}
\caption{Transfer learning for quick adaptation to new scenarios.}
\label{fig:transfer_learning}
\vspace{-0.1cm}
\end{figure*}

Conveniently however, rather than re-training a new RL-agent from scratch for every possible scenario, it is possible to leverage the principles of transfer learning \cite{pan2010survey} in order to speed-up the adaptation to new scenarios. Figure~\ref{fig:transfer_learning} shows an example of this. In this experiment, we pre-trained the proposed RL-agent on the base scenario for 40 episodes and then we transferred the learned parameters to the ``After event" scenario and kept training the agent on this new scenario. The left-hand side of Figure~\ref{fig:transfer_learning} shows the neural network loss, which, in the case of transfer learning, spikes at right after the change of scenario, but then quickly converges to similar loss values of those obtained by the RL agent trained from scratch on the ``After event" scenario. As for Figure~\ref{subfig:transferb}, it shows that both approaches are able to obtain similar results over a total of 10 runs, but the pre-trained agent is able to converge within just a couple of episodes. 

Lastly, by comparing the performance of the RL-based controller with fixed-times and the actuated controller (see Table~\ref{table:results1}), we can verify that the phase-selection approach is the one that provides the best results for both event scenarios. However, it should be noted that, once again, the difference in performance between the various control strategies varies significantly across different scenarios. 


\begin{table*}[th!]
\caption{Results over 10 runs for scenarios with different demand and supply (average and standard deviation in seconds)}
\label{table_example}
\begin{center}
\begin{tabular}{c|c|ccc|ccc}
\multirow{2}{*}{Scenario} & \multirow{2}{*}{Control Strategy} & \multicolumn{3}{c|}{Last 30 episodes} & \multicolumn{3}{c}{Last 10 episodes}\\
 &  & Travel Time & Delay & Stop Time & Travel Time & Delay & Stop Time\\
\hline
\multirow{4}{*}{Base} & Fixed Timings & 341.2 ($\pm$ 0.6) & 274.5 ($\pm$ 0.6) & 253.7 ($\pm$ 0.6) & 341.1 ($\pm$ 0.7) & 274.4 ($\pm$ 0.7) & 253.6 ($\pm$ 0.7)\\
& Actuated & 299.1 ($\pm$ 5.0) & 232.3 ($\pm$ 5.0) & 210.0 ($\pm$ 4.72) & 300.4 ($\pm$ 4.5) & 233.7 ($\pm$ 4.6) & 211.3 ($\pm$ 4.3)\\
& Time Extension & 348.4 ($\pm$ 23.9) & 281.7 ($\pm$ 23.9) & 264.0 ($\pm$ 22.8) & 331.9 ($\pm$ 6.6) & 265.2 ($\pm$ 6.6) & 248.4 ($\pm$ 6.4)\\
& Phase Selection & 200.4 ($\pm$ 5.5) & 133.7 ($\pm$ 5.5) & 117.8 ($\pm$ 5.4) & 202.0 ($\pm$ 5.3) & 135.3 ($\pm$ 5.3) & 119.3 ($\pm$ 5.1)\\
\hline
\multirow{4}{*}{Low demand} & Fixed Timings & 219.6 ($\pm$ 0.9) & 152.8 ($\pm$ 0.9) & 138.3 ($\pm$ 0.8) & 219.7 ($\pm$ 0.9) & 152.9 ($\pm$ 0.9) & 138.4 ($\pm$ 0.9)\\
& Actuated & 171.1 ($\pm$ 0.7) & 104.3 ($\pm$ 0.7) & 90.1 ($\pm$ 0.7) & 170.8 ($\pm$ 0.9) & 104.1 ($\pm$ 0.9) & 89.9 ($\pm$ 0.9)\\
& Time Extension & 276.9 ($\pm$ 18.1) & 210.2 ($\pm$ 18.1) & 195.4 ($\pm$ 18.1) & 279.6 ($\pm$ 16.7) & 212.9 ($\pm$ 16.8) & 198.0 ($\pm$ 16.8)\\
& Phase Selection & 165.4 ($\pm$ 3.0) & 98.6 ($\pm$ 3.0) & 84.5 ($\pm$ 3.0) & 168.0 ($\pm$ 2.8) & 101.2 ($\pm$ 2.8) & 87.1 ($\pm$ 2.7)\\
\hline
\multirow{4}{*}{Before event} & Fixed Timings & 379.3 ($\pm$ 5.7) & 312.5 ($\pm$ 5.7) & 289.5 ($\pm$ 5.5) & 379.9 ($\pm$ 4.9) & 313.2 ($\pm$ 4.9) & 290.1 ($\pm$ 4.6)\\
& Actuated & 303.9 ($\pm$ 1.4) & 237.1 ($\pm$ 1.4) & 214.5 ($\pm$ 1.3) & 304.2 ($\pm$ 1.2) & 237.5 ($\pm$ 1.2) & 214.9 ($\pm$ 1.2)\\
& Time Extension & 370.5 ($\pm$ 16.8) & 303.7 ($\pm$ 16.8) & 285.3 ($\pm$ 16.3) & 380.3 ($\pm$ 23.1) & 313.5 ($\pm$ 23.1) & 294.5 ($\pm$ 22.6)\\
& Phase Selection & 232.6 ($\pm$ 9.8) & 165.9 ($\pm$ 9.8) & 148.6 ($\pm$ 9.6) & 237.7 ($\pm$ 9.2) & 170.9 ($\pm$ 9.2) & 153.6 ($\pm$ 9.1)\\
\hline
\multirow{4}{*}{After event} & Fixed Timings & 414.7 ($\pm$ 3.3) & 347.9 ($\pm$ 3.3) & 323.2 ($\pm$ 3.2) & 413.6 ($\pm$ 2.9) & 346.8 ($\pm$ 2.9) & 322.2 ($\pm$ 2.7)\\
& Actuated & 319.5 ($\pm$ 4.0) & 252.7 ($\pm$ 4.0) & 229.4 ($\pm$ 3.8) & 319.4 ($\pm$ 4.7) & 252.7 ($\pm$ 4.7) & 229.3 ($\pm$ 4.5)\\
& Time Extension & 466.4 ($\pm$ 34.3) & 399.6 ($\pm$ 34.3) & 377.3 ($\pm$ 33.2) & 458.8 ($\pm$ 36.4) & 392.0 ($\pm$ 36.4) & 370.2 ($\pm$ 34.9)\\
& Phase Selection & 306.3 ($\pm$ 12.5) & 239.5 ($\pm$ 12.5) & 218.9 ($\pm$ 11.9) & 306.0 ($\pm$ 11.7) & 239.3 ($\pm$ 11.7) & 218.5 ($\pm$ 10.8)\\
\hline
\multirow{4}{*}{Incident A} & Fixed Timings & 311.7 ($\pm$ 6.5) & 245.0 ($\pm$ 6.5) & 225.2 ($\pm$ 6.2) & 311.0 ($\pm$ 5.1) & 244.4 ($\pm$ 5.1) & 224.5 ($\pm$ 4.9)\\
& Actuated & 226.9 ($\pm$ 3.0) & 160.3 ($\pm$ 3.0) & 142.0 ($\pm$ 2.8) & 227.3 ($\pm$ 3.6) & 160.6 ($\pm$ 3.6) & 142.4 ($\pm$ 3.4)\\
& Time Extension & 338.5 ($\pm$ 10.5) & 271.8 ($\pm$ 10.5) & 254.9 ($\pm$ 10.4) & 342.8 ($\pm$ 9.4) & 276.2 ($\pm$ 9.4) & 259.3 ($\pm$ 9.3)\\
& Phase Selection & 194.6 ($\pm$ 7.7) & 127.9 ($\pm$ 7.7) & 112.8 ($\pm$ 7.7) & 196.8 ($\pm$ 4.9) & 130.1 ($\pm$ 4.9) & 114.9 ($\pm$ 4.8)\\
\hline
\multirow{4}{*}{Incident B} & Fixed Timings & 394.6 ($\pm$ 12.8) & 327.9 ($\pm$ 12.8) & 300.2 ($\pm$ 12.3) & 401.8 ($\pm$ 8.6) & 335.1 ($\pm$ 8.6) & 307.2 ($\pm$ 8.2)\\
& Actuated & 321.8 ($\pm$ 19.0) & 255.2 ($\pm$ 18.9) & 229.6 ($\pm$ 18.5) & 324.0 ($\pm$ 13.3) & 257.4 ($\pm$ 13.3) & 231.9 ($\pm$ 13.0)\\
& Time Extension & 363.3 ($\pm$ 9.1) & 296.7 ($\pm$ 9.1) & 273.6 ($\pm$ 8.9) & 358.9 ($\pm$ 6.8) & 292.2 ($\pm$ 6.8) & 269.2 ($\pm$ 6.6)\\
& Phase Selection & 250.1 ($\pm$ 13.2) & 183.4 ($\pm$ 13.2) & 160.8 ($\pm$ 13.0) & 257.7 ($\pm$ 10.2) & 191.1 ($\pm$ 10.2) & 168.3 ($\pm$ 9.9)\\
\end{tabular}
\end{center}
\label{table:results1}
\vspace{-0.5cm}
\end{table*}

\subsection{Supply variation experiments}

\begin{figure}[th]
\centering
\includegraphics[scale=0.54,trim={0 0.2cm 0 1.3cm},clip]{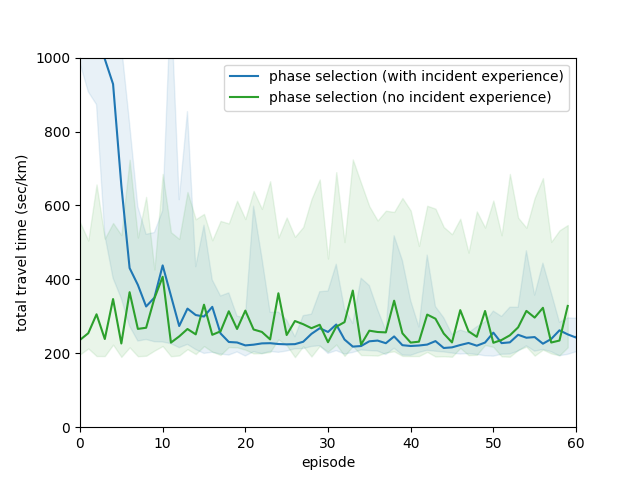}
\caption{Effect of incident experience for ``Incid. B" scenario.}
\label{fig:train_base_test_incident_real}
\vspace{-0.2cm}
\end{figure}

Next, we studied the effect of changes in supply, namely the effect of incidents. For this purpose, we considered the following two scenarios:

\textbf{Incident A:} This scenario tries to mimic the effect of a full road-block caused by an incident by setting the number of incoming vehicles from the east direction ($d=1$) to zero for a period of 2 hours during the middle of each episode. 

\textbf{Incident B:} In order to simulate the potential effect of an incident in a more realistic setting (\mbox{i.e.} using a speed reduction model, accounting for visibility, etc.), in this scenario we use the AIMSUN API to simulate an incident occupying 50m of left-most lanes from direction $d=1$ during the 2nd and 3rd hours of the total 4-hour simulation period.  

We begin by analyzing what would be the consequence of deploying on the ``Incident B" scenario a RL-based controller trained on the base scenario. Figure~\ref{fig:train_base_test_incident_real} shows the obtained results in comparison with the training curve for a RL-based controller that had access to the incident scenario during learning. Similarly to the results for the demand variation, we can clearly observe a very significance deterioration of the total travel times, with extreme cases almost tripling the travel time values. This not only has a very negative impact on the level-of-service of this particular intersection, but it could potentially have a very negative impact on the entire road network. 

In terms of performance of the different control strategies considered, Table~\ref{table:results1} clearly shows that, when trained with experience on incident scenarios, the deep RL approach based on phase-selection is robust and it is able to significantly outperform all the other methods.

\subsection{Sensor failure experiments}
\label{subsec:sensor_failures}

\begin{table*}[t]
\caption{Results over 10 runs for scenarios with sensor failures (average and standard deviation in seconds)}
\label{table:results_sensor}
\begin{center}
\begin{tabular}{c|c|c|ccc}
\multirow{2}{*}{Scenario} & \multirow{2}{*}{Control Strategy} & \multirow{2}{*}{Failure Type} & \multicolumn{3}{c}{Last 10 episodes}\\
 & & & Travel Time & Delay & Stop Time\\
\hline
\multirow{4}{*}{Base} & Fixed Timings & - & 341.1 ($\pm$ 0.7) & 274.4 ($\pm$ 0.7) & 253.6 ($\pm$ 0.7)\\
& Actuated & No Failures & 300.4 ($\pm$ 4.5) & 233.7 ($\pm$ 4.6) & 211.3 ($\pm$ 4.3)\\
& Time Extension & No Failures & 331.9 ($\pm$ 6.6) & 265.2 ($\pm$ 6.6) & 248.4 ($\pm$ 6.4)\\
& Phase Selection & No Failures & 202.0 ($\pm$ 5.3) & 135.3 ($\pm$ 5.3) & 119.3 ($\pm$ 5.1)\\
\hline
\multirow{2}{*}{Failure A} & Phase Selection (no dropout) & Before max() & 314.5 ($\pm$ 13.1) & 247.8 ($\pm$ 13.1) & 227.3 ($\pm$ 12.6)\\
& Phase Selection (w/dropout) & Before max() & 222.3 ($\pm$ 3.0) & 155.6 ($\pm$ 3.0) & 138.9 ($\pm$ 3.0)\\
\hline
\multirow{2}{*}{Failure A} & Phase Selection (no dropout) & After max() & 355.0 ($\pm$ 18.2) & 288.2 ($\pm$ 18.1) & 267.4 ($\pm$ 17.6)\\
& Phase Selection (w/dropout) & After max() & 286.3 ($\pm$ 13.7) & 219.6 ($\pm$ 13.7) & 201.3 ($\pm$ 13.4)\\
\hline
\multirow{2}{*}{Failure B} & Phase Selection (no dropout) & Before max() & 258.6 ($\pm$ 18.2) & 191.8 ($\pm$ 18.2) & 173.9 ($\pm$ 17.7)\\
& Phase Selection (w/dropout) & Before max() & 233.4 ($\pm$ 7.5) & 166.6 ($\pm$ 7.6) & 149.4 ($\pm$ 7.5)\\
\hline
\multirow{2}{*}{Failure B} & Phase Selection (no dropout) & After max() & 436.8 ($\pm$ 52.5) & 370.0 ($\pm$ 52.5) & 348.7 ($\pm$ 52.0)\\
& Phase Selection (w/dropout) & After max() & 280.8 ($\pm$ 31.0) & 214.1 ($\pm$ 31.0) & 195.6 ($\pm$ 30.5)\\
\hline
\multirow{2}{*}{Failure C} & Phase Selection (no dropout) & Before max() & 222.9 ($\pm$ 3.2) & 156.1 ($\pm$ 3.2) & 138.6 ($\pm$ 3.1)\\
& Phase Selection (w/dropout) & Before max() & 214.6 ($\pm$ 4.6) & 147.8 ($\pm$ 4.6) & 131.6 ($\pm$ 4.5)\\
\hline
\multirow{2}{*}{Failure C} & Phase Selection (no dropout) & After max() & 241.0 ($\pm$ 4.2) & 174.2 ($\pm$ 4.2) & 156.8 ($\pm$ 4.0)\\
& Phase Selection (w/dropout) & After max() & 234.3 ($\pm$ 8.8) & 167.6 ($\pm$ 8.8) & 149.4 ($\pm$ 8.1)\\
\end{tabular}
\end{center}
\vspace{-0.5cm}
\end{table*}

Lastly, we turn our focus to robustness to sensor failures and other types of issues with sensory inputs to the deep RL agent, by considered in the following 3 scenarios: 

\textbf{Failure A:} In order simulate realistic sensor failure patterns, in which the sensors fail in ``bursts", we implemented a 2-state Markov process for each lane sensor/group of sensors. Each sensor then alternates between two different states - OK and failure - with some probability. We further consider two failure types: before and after the max aggregation in Eq.~\ref{eq:state}. While the former could represent a problem with the individual sensors (\mbox{e.g.} loop detectors), the latter corresponds to the extreme setting in which all the input sensory data associated with a given phase is faulty. We assume faulty sensors to provide a value of zero (\mbox{e.g.} a queue length of zero). In this particular scenario, the probability of entering a failure state is set to 1\%, while the probability of recovery is set to 5\%. This corresponds to an overall fraction of failures of approximately $16\%$, and an average failure length of 20 time-steps. Kindly note that this allows for multiple sensors (eventually even all of them) to be failing simultaneously. The demand used is the same as in the base scenario.

\textbf{Failure B:} This scenario is similar to ``Failure A", but the probability of entering a failure state and the probability of recovery are set to 0.1\% and 0.5\%, respectively. Therefore, while the overall fraction of failures is the same, failures last 10 times longer. 

\textbf{Failure C:} In this scenario, a single input sensor/group of sensors (depending on the ``failure type") is set to zero for a 2-hour period in the middle of each episode. 

\begin{figure}[t]
\centering
\includegraphics[scale=0.54,trim={0 0.2cm 0 1.3cm},clip]{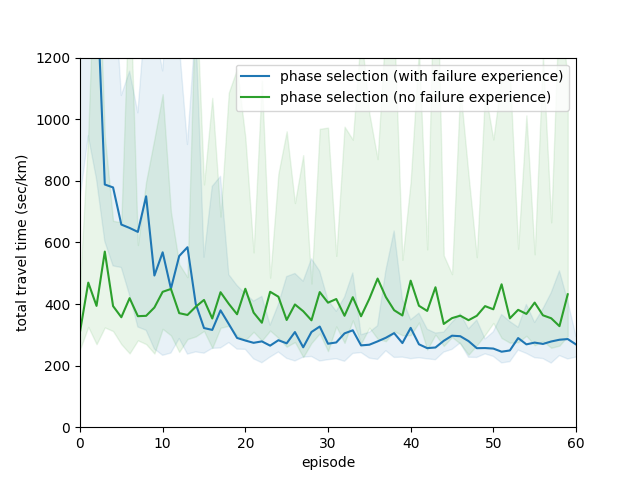}
\caption{Effect of failure experience for ``Failure A" scenario.}
\label{fig:train_base_test_sensor1}
\vspace{-0.3cm}
\end{figure}

As with the previous experiments, we begin by analyzing the impact of input failures (scenario ``Failure A") on a RL controller trained without experiencing failures, in contrast with a RL controller that experienced failures. As Fig.~\ref{fig:train_base_test_sensor1} shows, not accounting for sensor failures during learning leads to very unreliable controllers that are very sensitive even to small changes in the inputs, which in turn can result in extremely poor performance. Interestingly, the proposed RL approach is able to learn to be robust to such failures.

In order to further understand the robustness properties of the proposed RL approach, Table~\ref{table:results_sensor} shows the performance of the phase-selection version for the different failure scenarios described above, in comparison to the original results for the base scenario, \mbox{i.e.} without failures. As expected, we can verify that introducing sensor failures leads to generally worse performance of the RL approach. Also, naturally, having failures after the $\max(\dots)$ aggregation operation when computing the input state for the RL agent in Eq.~\ref{eq:state} leads to significantly worse performance, which confirms our intuition that the max aggregation serves as an important first step towards building robust RL-based controllers for traffic signal control. 

But perhaps most interestingly is observing the impact of Dropout for building robust controllers. As Table~\ref{table:results_sensor} shows, not including Dropout in the neural network leads to RL controllers that perform substantially worse when faced with scenarios with sensor failures. Please note that the same does not happen when considering only scenarios without failures, for which the versions with and without Dropout tend to perform similarly. This can be verified, for example, in the results for the ``Failure C" scenario for failures ``Before max()", in which case the $\max(\dots)$ aggregation mitigates most of the effect of the failure (since it only concerns a unique lane), thus resulting only in a small difference in performance between the with/without Dropout versions. 

Lastly, from a overall analysis of the results in Table~\ref{table:results_sensor}, we can observe that, when trained in scenarios with sensor failures, the RL approach can be quite robust to the widely common issue of malfunctioning sensors (\mbox{e.g.} inductive loop detectors or cameras). Although the average performance naturally degrades slightly, we can observe that, for all scenarios considered, the proposed deep RL approach is still able to significantly outperform fixed-time and actuated controllers that actually observe the true sensor data without any failures.

\section{CONCLUSIONS AND FUTURE WORK}

This paper investigated the impact of demand surges, incidents (supply) and sensor failures on traffic signal controllers based on deep reinforcement learning, and extracted some initial, but important, insights towards the development of deep RL approaches that are robust to many of the issues that arise in practice when being deployed in lively and highly dynamic urban environments. In order to do so, we developed a callback-based RL framework (CAREL) and integrated it with the traffic simulation software AIMSUN for testing different traffic control strategies in different scenarios that we developed and made publicly available, so that they can serve as a benchmark to foster the development of new robust RL approaches for traffic signal control in highly realistic and dynamic scenarios. 
Our empirical evidence shows that, in order for deep RL controllers to be robust to demand surges, incidents and sensor failures, they need to experience these scenarios during training - something that can be efficiently achieved by exploiting transfer learning. This, together with the use of Dropout and an extended state representation, allows for the development of more robust RL agents. 

Despite providing an important step towards the development of robust RL-based controllers, this paper also leaves several research questions open for future work. Our ongoing work is exploring how these findings generalize to scenarios with a realistic road network with multiple intersections, and how deep RL approaches could be efficiently trained in order to be robust in more complex settings. Future work will also focus on studying how robust RL approaches based on the theory of $H_\infty$ control \cite{morimoto2005robust}, such as Robust Adversarial RL \cite{pinto2017robust}, can be used to further increase the robustness of the policies learned by the RL-based traffic signal controllers. 

\addtolength{\textheight}{-12cm}   


\section*{ACKNOWLEDGMENT}

We thank Transport Simulation Systems (TSS) for supporting this research with AIMSUN licenses and feedback.



\bibliographystyle{IEEEtran}
\bibliography{IEEEabrv,robust-deep-rl-traffic-lights}

\end{document}